\documentclass{article}
\pdfoutput=1

\PassOptionsToPackage{numbers, compress}{natbib}

\usepackage[preprint]{neurips_2024}

\usepackage[utf8]{inputenc} 
\usepackage[T1]{fontenc}    
\usepackage{hyperref}       
\usepackage{url}            
\usepackage{booktabs}       
\usepackage{amsfonts}       
\usepackage{nicefrac}       
\usepackage{microtype}      
\usepackage{xcolor}         
\usepackage[colorinlistoftodos]{todonotes}
\usepackage{amsmath}
\usepackage{dsfont}
\usepackage{bm}
\usepackage{float}
\usepackage{multirow}
\usepackage{algorithm,algpseudocode}

\DeclareMathOperator*{\argmin}{arg\,min}

\title{Multi-fidelity Reinforcement Learning Control for Complex Dynamical Systems}

\author{%
  Luning Sun\thanks{Equal contribution.} \\
  Lawrence Livermore National Lab\\
  Livermore, CA 94550 \\
  \texttt{sun42@llnl.gov} \\
  \And
  Xin-Yang Liu  \\
  University of Notre Dame\\
  Notre Dame, IN 46556 \\
  \texttt{xliu28@nd.edu} \\
  \And
  Siyan Zhao \\
  University of California, Los Angeles\\
  Los Angeles, CA 90095\\
\texttt{siyanz@cs.ucla.edu}
\And
  Aditya Grover \\
  University of California, Los Angeles\\
  Los Angeles, CA 90095 \\
  \texttt{adityag@cs.ucla.edu} \\
\And
  Jian-Xun Wang \\
  University of Notre Dame\\
  Notre Dame, IN 46556 \\
  \texttt{jwang33@nd.edu} \\
  \And
  Jayaraman J. Thiagarajan\\
  Lawrence Livermore National Lab\\
  Livermore, CA 94550 \\
  \texttt{jjayaram@llnl.gov}
}

\begin{document}

\maketitle

\begin{abstract}
  Controlling instabilities in complex dynamical systems is challenging in scientific and engineering applications. Deep reinforcement learning (DRL) has seen promising results for applications in different scientific applications. The many-query nature of control tasks requires multiple interactions with real environments of the underlying physics. However, it is usually sparse to collect from the experiments or expensive to simulate for complex dynamics. Alternatively, controlling surrogate modeling could mitigate the computational cost issue. However, a fast and accurate learning-based model by offline training makes it very hard to get accurate pointwise dynamics when the dynamics are chaotic. To bridge this gap, the current work proposes a multi-fidelity reinforcement learning (MFRL) framework that leverages differentiable hybrid models for control tasks, where a physics-based hybrid model is corrected by limited high-fidelity data. We also proposed a spectrum-based reward function for RL learning. The effect of the proposed framework is demonstrated on two complex dynamics in physics. The statistics of the MFRL control result match that computed from many-query evaluations of the high-fidelity environments and outperform other SOTA baselines. 


\end{abstract}

\section{Introduction}
\label{sec:intro}
Instabilities are universal in complex dynamical systems, and controlling their evolution has great value in science and engineering. For example, laser-plasma instability~\cite{montgomery2016two,ludwig2018enhancement,mariscal2023application,anirudh20232022,kline2005observation} is usually an undesirable phenomenon in inertial confinement fusion processes, where a great of input laser energy is dissipated during this process, increasing difficulties in getting power net gain. In fluid dynamics, Plateau-Rayleigh instabilities describe the process of falling liquid film breaking into small drops under a given threshold. Understanding when and why it happens also has great value in the field of multi-phase flow~\cite{belus2019exploiting,martin2023physics}. Furthermore, Rayleigh-Benard instabilities, caused by the buoyancy effect, are a widely investigated topic in thermal-driven flows~\cite{vignon2023effective}. Finally, Kevin-Helmholtz instabilities are formulated by the velocity shear~\cite{list2022learned} and can be observed in the atmosphere of planets.

Given their broad utility, methods for controlling or mitigating such instabilities have gained significant research interest. In particular, deep reinforcement learning (DRL) has emerged as an important approach in different scientific applications. For example, in magnetic confinement nuclear fusion, DRL has witnessed success in controlling tokamak core plasma~\cite{degrave2022magnetic} and avoiding the plasma tearing instabilities~\cite{seo2024avoiding}. There is also recent research about controlling instabilities and chaos in fluid dynamics~\cite{belus2019exploiting,bucci2019control,vashishtha2020restoring}. These DRL algorithms typically learn policies through multiple interactions with the environments. Given that high-fidelity computational models in many scientific applications are prohibitively expensive, and that deep reinforcement learning (DRL) requires a large number of evaluations of these models to infer a reasonable policy, the feasibility of this approach is a practical concern. Using low-fidelity approximations of the computational model is an alternative solution. However, these approximations often rely on strong assumptions that may not hold in practice, potentially leading to non-physical results.

To mitigate the computational burden, while also not sacrificing the performance significantly, we propose a multi-fidelity reinforcement learning framework that controls the high-fidelity model indirectly through a low-fidelity model with learnable correction terms. The key novelty of this work is twofold. First, we integrate additional domain knowledge of instabilities into the DRL framework. Second, we introduce learnable correction terms in a differentiable model to bridge the gap between models of varying fidelities, thereby reducing the number of evaluations required for high-fidelity models. Finally, we benchmark our framework's ability to mitigate instabilities in complex dynamical systems through two use-cases. We introduced various physical criteria to evaluate the soundness of learning performance. The result shows that the statistics of the controlled result by our framework could match the result computed from many-query evaluations of the high-fidelity environments. The performance of the proposed model also outperforms other baselines. 


\section{Related Work}
\label{sec:Related works}
\paragraph{Instabilities in Dynamical Systems}
Instabilities in dynamical systems represent a fascinating research area. They occur when small perturbations amplify with time, leading to unpredictable behavior that evolves with time. One area in which instabilities play an important role is laser-plasma interaction (LPI). Key instabilities matter are stimulated Raman and Brillouin scattering (SRS/SBS)~\cite{kruer2019physics,forslund1975theory}. Another area is fluid instabilities that are ubiquitously seen in nature, like turbulence, vortex formulation, and chaotic mixing.  Classical phenomenons, such as Rayleigh-Bernard convection, and Kelvin-Helmholtz~\cite{chandrasekhar2013hydrodynamic,drazin2004hydrodynamic} have been investigated for decades by experiments and simulations. However, in the deep learning era, huge gaps exist in integrating learning techniques into the framework to control/optimize the instabilities and our work aims to bridge this gap.

\paragraph{Deep Reinforcement Learning for Control}
Deep reinforcement learning (DRL) has demonstrated success in learning optimal policies through trial-and-error interactions with environments, achieving remarkable results in simulated environments like Minecraft and Atari games~\cite{fujimoto2018addressing, fan2022minedojo}. Recently, there have been attempts to apply RL to real-world physics control tasks, such as robotic fish~\cite{cui2024enhancing}, turbulent flows~\cite{bae2022scientific}, and falling liquid films~\cite{belus2019exploiting}, where traditional control methods are limited. However, these DRL methods require querying the environment to obtain a sufficient number of transitions for learning the policy, and in real-world, access to a high-fidelity simulator or controller can be expensive and sometimes infeasible. To address the challenge of sample efficiency, one approach is to use model-based RL, which learns a model of the transition dynamics to simulate additional data~\cite{liu2021physics, zolman2024sindy}. Leveraging simulators of different fidelity levels to train policies can also improve data efficiency, with techniques like Gaussian Processes and control variates used to help reduce variance in state-action value function~\cite{suryan2017multi, khairy2022multifidelity}. Additionally, transfer learning across multi-fidelity data~\cite{bhola2023multi} or similar RL environments~\cite{zhao2024decision} enables knowledge transfer for better data efficiency.


\paragraph{Multi-fidelity Modeling and Differentiable Programming}
The multi-fidelity modeling framework is beneficial for many-query applications like uncertainty quantification and design optimization. This framework usually delegates the majority of the computational budget to low-fidelity models whose output could be corrected by a small number of labeled high-fidelity data. The recent literature has invested in various settings for multi-fidelity modeling. The quantities of interest (QOI) could be either integrated properties~\cite{zhang2021multi, lamberti2021multi} (e.g., draft/lift coefficient) or full field state variables~\cite{mondal2022multi,li2023multi}. The correction terms can also be higher-order POD modes~\cite{conti2024multi,pawar2021model} or missing closure terms~\cite{sen2018evaluation, liu2024multi}. Recently, there are also probabilistic multi-fidelity generative models that produce high-fidelity realizations by conditioning on corresponding low-fidelity data~\cite{geneva2020multi, wan2024debias,gao2023bayesian}.
To the author's knowledge, there is limited research on controlling the instabilities in multi-fidelity formulations. Sahil et al applied the multi-fidelity on a steady-state shape optimization problem~\cite{bhola2023multi}. The knowledge between different fidelity models is shared by the parameter of the policy network by transfer learning.

\section{Proposed Approach}
\label{sec:Method}
\begin{figure}[H]
   \centering
   \includegraphics[width=0.72\textwidth]{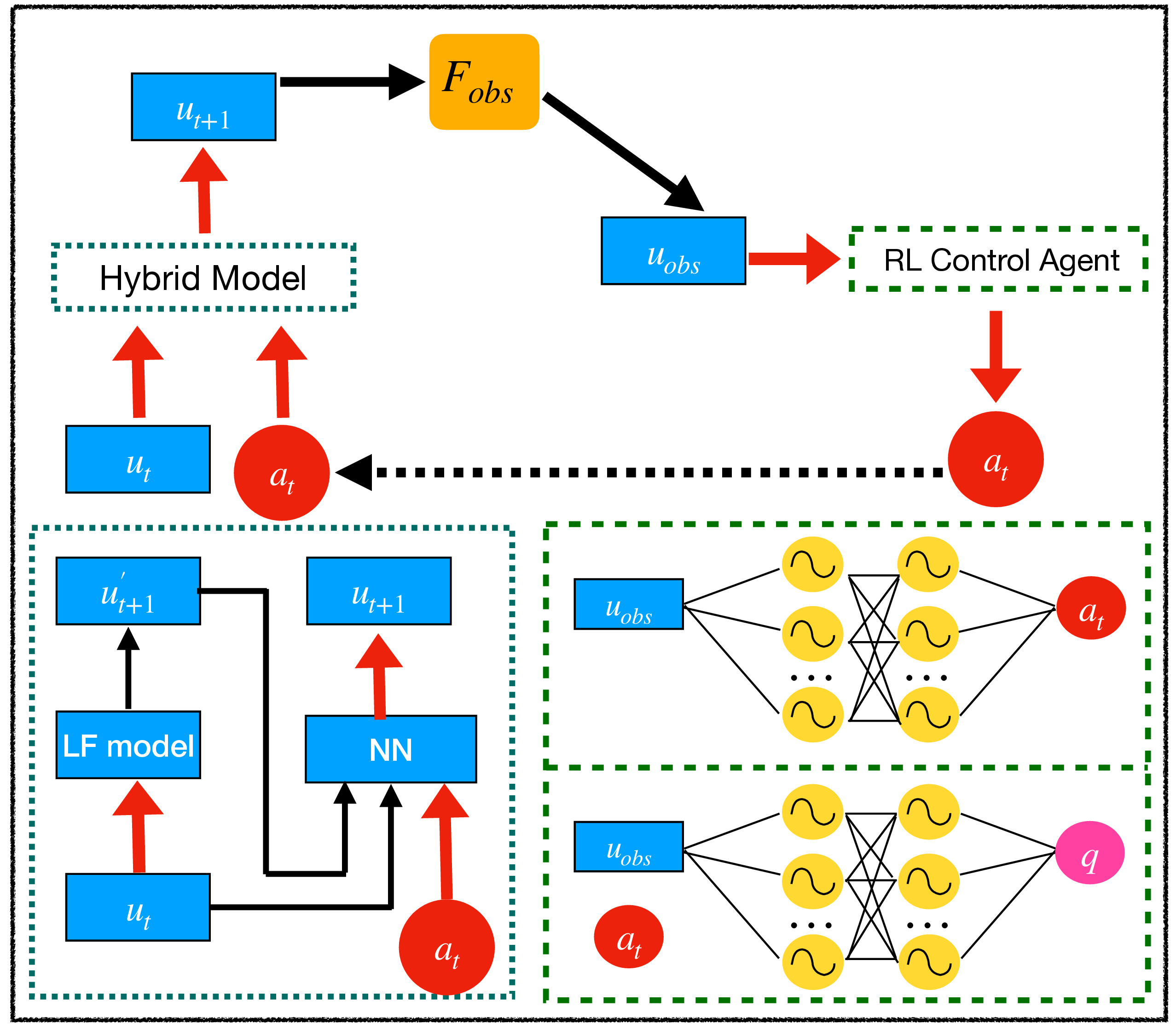}
   \caption{Overview of hybrid reinforcement learning control framework. Sketches in dotted green boxes are the detailed forward functions of hybrid models and RL agents, respectively.}
   \label{fig:overview}
 \end{figure}

In this section, we describe our multi-fidelity DRL approach and also discuss the implementation details. The focus of this paper is on using actuators to control dynamical systems governed by partial differential equations (PDEs), which can be formulated as:
\begin{equation}
\frac{\boldsymbol{du}}{{dt}} = F(\boldsymbol{u},\boldsymbol{a};\boldsymbol{\mu}),
\end{equation}
where $\boldsymbol{u}(\boldsymbol{x},t) \in \mathbb{R}^{d_u}$ denote the state variables defined in a spatial domain $\Omega$ and a temporal domain $t \in [0,T]$, $\boldsymbol{a}(\boldsymbol{x},t)$ correspond to the action variable for control tasks, and $F(\boldsymbol{\cdot};\boldsymbol{\mu})$ is the differential operator parameterized by $\boldsymbol{\mu}$. It is typical to assume the dynamical system to be a Markov Decision Process (MDP), and hence the discretized system can be expressed as
\begin{equation}
\boldsymbol{u}_{t+1} = \hat{F}(\boldsymbol{u}_t, \boldsymbol{a}_t; \boldsymbol{\mu}),
\end{equation}
where $\hat{F}$ is the discretized form of the operator $F$. The ultimate goal of the control tasks is to determine a policy $\pi$ that maximizes the expected return $R(\pi)$, defined as follows:
\begin{equation}
R(\pi) = \int_{0}^{T}\mathbb{E}[r(\boldsymbol{u}_{t})]dt,
\end{equation}
where $r(\cdot)$ denotes the reward function. Given this formulation, DRL approaches can be utilized to identify the optimal actions. An overview of the proposed hybrid reinforcement learning approach for multi-fidelity datasets is shown in Fig~\ref{fig:overview}. Our approach is comprised of the following steps: Firstly, a hybrid model is trained offline using the multi-fidelity datasets, which are illustrated in gray boxes and discussed in Sec.~\ref{subsec:diff_env}. The hybrid model is then used as the surrogate environment for reinforcement learning, which is introduced in Sec.~\ref{subsec:RL}. Specifically, we propose spectrum-based reward functions for controlling complex chaotic systems. Furthermore, we employ a stochastic weight averaging strategy during training to obtain more stable reward curves. These implementation details are discussed in detail in Sec.~\ref{subsec:RLdetails}. Finally, we evaluate our framework on two types of complex dynamical systems, namely plasma (in Sec.~\ref{sec:result_plasma}), and fluid (in Sec.~\ref{sec:result_fluid}).
\subsection{A differentiable hybrid environment}
\label{subsec:diff_env}

For a physical process governed by PDEs, there usually exist models of different fidelities. In this context, low fidelity (LF) often corresponds to computational models that do not directly solve the original PDE, or do not solve them at proper resolution. Instead, numerical models are usually leveraged to compensate for the under-resolved dynamics to maintain an acceptable accuracy, while significantly reducing the computational cost. 
On the other hand, high fidelity (HF) models provide an accurate characterization compared to LF models by solving the original PDEs on high-resolution grids and usually use higher order of accuracy numerical schemes. 
However, in the context of RL-based control, the high computational cost of HF environments can be prohibitive in practice, due to the trial-and-error nature of RL approaches. While LF models incur significantly lower costs, the instability of the system magnifies even the small discrepancies between the LF models and the underlying dynamics, thus leading to poor control performance of the RL agents, or even complete failure in obtaining a converged control policy. 
To bridge this gap, we propose to use a hybrid environment for control tasks based on multi-fidelity models. With RL methods, it is assumed that the actions in control tasks only depend on the previous control step state $u_{t}$, and that the control policy learned in a hybrid environment is expected to generalize well to the original dynamic system. Since every environment in our hybrid setup satisfies the MDP assumption, we formulate our differentiable hybrid model as a predictor-corrector, with a ResNet~\cite{butcher2016numerical,he2016deep} backbone, as follows.
\begin{align}
\label{eq:hybrid}
\boldsymbol{u}^{C}_{t+1} &= \hat{F}^{L}(\boldsymbol{u}_t,\boldsymbol{a}_t;\boldsymbol{\mu})+f_{\theta}(\boldsymbol{u}^{C}_{t}, \hat{F}^{L}(\boldsymbol{u}_t,\boldsymbol{a}_t;\boldsymbol{\mu}),\boldsymbol{a_t};\boldsymbol{\mu},\boldsymbol{\theta}),
\end{align}
where the superscripts $L, C, H$ denote the low fidelity, hybrid corrected, and high fidelity models respectively. Further, $f_{\theta}$ is the neural network (NN) correction model parameterized by $\theta$, which leverages the SIREN architecture~\cite{sitzmann2020implicit} for its ability to represent complex natural signals and their derivatives through the use of periodic activation functions. The mathematical forward function can be expressed as
\begin{align}
\label{eq:SIREN}
\begin{split}
\bold{SIREN}(\boldsymbol{x}) &= \boldsymbol{W}_p(\eta_{p-1}\circ\eta_{p-2}\circ\cdots \circ \eta_1)(\omega_0\boldsymbol{W}_0\boldsymbol{x}+\boldsymbol{B}_0)+\boldsymbol{B}_p\\
\eta_m(o_{m-1})&= \text{sin}(\boldsymbol{W}_m o_{m-1}+\boldsymbol{B}_m),
\end{split}
\end{align}
Where $o_{m-1}$ is the output of $(m-1)_{th}$ layer and $m \in [1,p]$. $\boldsymbol{W}$ and $\boldsymbol{B}$ are trainable weights and biases, and $\omega_0$ is the hyper-parameter to initialize the input signal frequency and follows the recommendation in~\cite{sitzmann2020implicit}. After obtaining the corrected result from the forward pass, the NN parameters can be updated using the following loss function.
\begin{align}
\label{eq:NNloss}
\begin{split}
L &= \sum_{i=0}^{T-1}\sum_{j=1}^{N}\frac{(\boldsymbol{u}^{C}_{i+1}-{\bar{\boldsymbol{u}}}^{H}_{i+1})^2}{NT},\\
 \bar{\boldsymbol{u}}^{H} &= G(\boldsymbol{u}^{H}).
\end{split}
\end{align}
Here $G$ is the projection function to align the high fidelity model result with the hybrid correction model, and $\bar{\boldsymbol{u}}^{H}$ denotes the projected high fidelity result. Note that the choice of projection function $G$ is based on the specific problem. For the plasma physics case, $G_2$ is the spatial average function $\frac{1}{N_x}\int\square dx$, while for the Burgers' turbulence case it is a low-pass filter.

Time evolution of the state variables requires evaluation of the NN correction model at every step. Directly training on the complete trajectory and performing gradient propagation will incur significant time and memory costs. Through the MDP assumption, we are able to split the whole trajectory into multiple overlapping windows, and then train the model on overlapping windows to save computation costs and reduce memory requirements. 

\subsection{Online reinforcement control}
\label{subsec:RL}
We will first introduce the most relevant components in any DRL algorithm, namely value function and policy function. A value function $v(\boldsymbol{u})$ for a state variable $\boldsymbol{u}$ is referred to as the state-value function, and similarly the value function $q(\boldsymbol{u}, \boldsymbol{a})$ denotes the action-value function. Mathematically, they are defined as 
\begin{align}
\begin{split}
v(\boldsymbol{u}) &= \mathbb{E}\bigg[\sum_{k=1}^{\infty}\gamma^{k}r(\boldsymbol{u}_{t+k})|\boldsymbol{u}_t=\boldsymbol{u}\bigg],\\
q(\boldsymbol{u},\boldsymbol{a}) &= \mathbb{E}\bigg[\sum_{k=1}^{\infty}\gamma^{k}r(\boldsymbol{u}_{t+k})|\boldsymbol{u}_t=\boldsymbol{u},\boldsymbol{a}_t=\boldsymbol{a}\bigg],
\end{split}
\end{align}
where $\gamma \in (0, 1]$ is the discount factor. A policy function is used to map from states to action either through a deterministic $\pi(\boldsymbol{u})$ or probabilitistically via $\pi(\boldsymbol{a}|\boldsymbol{u})$. In this work, we used actor-critic based method to optimize DRL agent. In the actor component, the policy function $\tilde{\pi}(\boldsymbol{u}, \boldsymbol{\theta}_\pi)$ is parameterized by a deep neural network and is updated using the temporal difference (TD) error provided by the critic. In the critic component, the value-action function $\tilde{q} (\boldsymbol{u}, \boldsymbol{a}; \boldsymbol{\theta}_q)$, also parameterized by a neural network, is updated based on the same temporal difference error. Using samples from the environment, the policy network is then updated as
\begin{equation}
\boldsymbol{\theta}_{\pi}^{k+1} = \boldsymbol{\theta}_{\pi}^{k}
+\alpha_{\pi}\nabla_{\boldsymbol{\theta_{\pi}}} J(\boldsymbol{\theta}_{\pi}^{k}),
\end{equation}
where $\alpha_{\pi}$ is the learning rate and $\nabla_{\boldsymbol{\theta}_{\pi}}J(\boldsymbol{\theta}_{\pi})$ is the policy gradient calculated as follows:
\begin{equation}
\nabla_{\boldsymbol{\theta}_{\pi}}J(\boldsymbol{\theta}_{\pi}^{k}) = \mathbb{E}\bigg[\sum_{t=0}^{T}\nabla_{\boldsymbol{\theta}_\pi}\text{log}\tilde{\pi}(\boldsymbol{u}_t;\boldsymbol{\theta}_{\pi}^{k})\cdot \tilde{q}(\boldsymbol{u}_t,\boldsymbol{a}_t;\boldsymbol{\theta}_{q}^{k})\bigg].
\end{equation}
Here, $\boldsymbol{\theta}_q^{k}$ denotes the parameters for critic network, which are in turn updated as
\begin{equation}
\boldsymbol{\theta}_q^{k} = \argmin_{\boldsymbol{\theta}_q} \bigg\|q_t^{\prime}-\tilde{q}(\boldsymbol{u}_t,\boldsymbol{a}_t;\boldsymbol{\theta}_q) \bigg\|_{L_2},
\end{equation}
where $q_t^{\prime}$ is calculated by the Bellman equation:
\begin{equation}
q_t^{\prime} = r_t+\gamma\tilde{q}\bigg(\boldsymbol{u}_{t+1}, \tilde{\pi}(\boldsymbol{u}_{t+1};\boldsymbol{\theta}_\pi);\boldsymbol{\theta}_q\bigg)
\end{equation}
\subsection{Implementation Details}
\label{subsec:RLdetails}
We now provide additional details on the implementation of our RL-based control powered by the hybrid multi-fidelity environment.
\subsubsection{Spectrum-based reward function}
The existing literature on control tasks in physical systems relies heavily on appropriate reward functions defined directly in the state space. While this approach is straightforward, we argue that it is not optimal for controlling instabilities. The reason is that the chaotic behavior of such physical systems tends to amplify any roll-out error incurred in the intermediate steps, eventually leading to an entirely different trajectory, which can still belong to one of the tractors of the chaotic systems. On the other hand, in the physics literature, there is a long-standing tradition of using spectral analysis tools for characterizing chaotic systems~\cite{abarbanel1993analysis}. In addition to providing rich mathematical tools, it is flexible to support the hypothesis of constructing multiple physical realizations corresponding to a single given spectrum. Building upon this framework, we propose to use the following spectral domain metrics for our optimization: $\ell_1$, $\ell_2$ and $\Delta_{\omega}$:
\begin{align}
\label{eq:sepctrum_matrix}
\begin{split}
\ell_1 &= \sum_{i=1}^{n_p} E_i \cdot \mathds{1}_{\{\text{peak}\}}(f_i),\\
\ell_2 &= \sum_{i=1}^{n_p} (E_i \cdot \mathds{1}_{\{\text{peak}\}}(f_i))^2,\\
\Delta_{\omega} &= \arg \max_{i \in \mathcal{P}} (f_i) - \arg \min_{i \in \mathcal{P}} (f_i),\\
\end{split}
\end{align}
where $E(f)$ represents the spectral energy of the signal at frequency $f$, $\mathds{1}_{\{\text{peak}\}}(i)$ is an indicator function that is 1 if $f_i$ is a peak location, 0 otherwise. $\mathcal{P}$ is the set of peak locations in the spectrum and \(f_i\) denotes the position of the $i$-th peak frequency. For fluid dynamics cases, we calculated the turbulence kinetic energy based on $e = 1/2(\boldsymbol{u}^{\prime})^2$, where $\boldsymbol{u}^{\prime}$ is the turbulence fluctuation and then computed the power spectrum density (PSD) using the Welch's method.

\subsubsection{Stochastic weight averaging training}

Stochastic Weight Averaging (SWA), introduced by Izmailov et al.~\cite{izmailov2018averaging} in 2018, is a novel training strategy designed to enhance generalization in both supervised and semi-supervised learning settings. This technique improves model performance by averaging model weights obtained at different stages of training using an SGD-like approach. Specifically, weights are collected after each training epoch to avoid convergence to a singular solution and promote continued exploration of optimal network configurations. This method has shown promise in navigating the solution space more effectively, targeting areas where networks perform well. Technically the averaging is done by calculating moving average along the SWA udpate steps:
\begin{align}
\begin{split}
\theta_{\text{SWA}} &= \frac{n_{\text{SWA}}\cdot\theta_{\text{SWA}}+\omega}{n_{\text{SWA}+1}}\\
n_{\text{SWA}} &= n_{\text{SWA}}+1
\end{split}
\end{align}
where $\theta_{\text{SWA}}$ is the averaged parameters, $\omega$ is the weight collected along training and $n_{\text{SWA}}$ is the number of SWA iterations. Finally, we use the Twin Delayed DDPG (TD3) algorithm for RL training. 

\section{Experiments and Result}
\label{sec:Result}
\subsection{Plasma Instabilities}
\label{sec:result_plasma}
We first demonstrate the capability of multi-fidelity RL to control a plasma system, which is governed by the following PDE~\cite{kono2010nonlinear}:
\begin{equation}
\begin{split}
\frac{\partial u_0}{\partial t} +{V_0\frac{\partial{u_0}}{\partial{x}}} &= -u_1u_2+S_0(\boldsymbol{a},t)\\
\frac{\partial u_1}{\partial t}-{V_1\frac{\partial{u_1}}{\partial{x}}} &= u_0u_2^*\\
\frac{\partial u_2}{\partial t}+{V_2\frac{\partial{u_2}}{\partial{x}}} &= \beta_0^2u_0u_1^*+j\sigma|u_2|^2u_2-\nu u_2,
\end{split}
\label{eq:plasma_pde}
\end{equation}
where $\boldsymbol{u} (x, t) = (u_1, u_2, u_3) \in \mathbb{C}^3$, $i \in \{0,1,2\}$ are the state variables. $S_0(\boldsymbol{a},t)$ is the time-dependent source term, which will be controlled by the RL agent, and $\beta_0$ is the constant that governs the energy transfer between different state variables. The asterisk ($^*$) represents the conjugate of the complex variables, where $j$ represents the imaginary unit ($j^2 = -1$). $\sigma$ and $\nu$ are coefficients for the non-linear term and the damping coefficients, respectively. 

The control objective here is to learn a policy for $S_0(\boldsymbol{a},t)$ where $S_0$ is constant over space. The goal is to generate the optimized response to mitigate the Stimulated Raman Scattering (SRS) in terms of the statistic defined in Eq.~\ref{eq:sepctrum_matrix}. Specifically, the reward function is defined as $r = \ell_1+w_1\cdot \ell_2+w_2\cdot\Delta_{\omega}$, where $w_1$ and $w_2$ are hyperparameters. The source term is parameterized using the Butterworth filter function $S_0 = a_{(1)}(t)\cdot (1+(t-\mu)/a_{(2)}(t))^{-n}$, where $a_i(t)_{\{i=1,2\}}\in [0.5,2]\times[0.5,1.2]$ are the control parameters. The high-fidelity environment is simulated using Eq.~\ref{eq:plasma_pde}. The spatial gradient term $\frac{\partial{}}{\partial{x}}$ is discretized using a second-order central difference scheme with a periodic boundary condition applied. The time integration is done by the fourth-order Runge-Kutta scheme with a time-stepping size of 0.001. And the spatial domain is defined on $[0, 10]$ discretized using a mesh of $100$ grid points. 
\subsubsection{A hybrid differentiable surrogate model for plasma}
To reduce the computational cost of simulating the plasma environment, we leverage the simplified ODE system with a DNN as a surrogate controlling environment instead of directly resolving the original PDE. The ODE system is governed by the following equation:
\begin{equation}
\label{eq:HyrbidSRS}
\begin{split}
    \frac{d \Tilde{u}_0}{dt} &= -\Tilde{u}_1\Tilde{u}_2+S_0(\boldsymbol{a},t) + N_1\\
\frac{d\Tilde{u}_1}{dt} &= \Tilde{u}_0\Tilde{u}_2^* + N_2\\
\frac{d\Tilde{u}_2}{dt} &= \beta_0^2\Tilde{u}_0\Tilde{u}_1^*+j\sigma|\Tilde{u}_2|^2\Tilde{u}_2-\nu \Tilde{u}_2 + N_3
\end{split}
\end{equation}
where $\Tilde{u}_i(t) = \frac{1}{N_x}\int u_i(t) dx$ is the state variable of the simplified ODE system. $\boldsymbol{N} = ( N_1, N_2, N_3) \in \mathbb{R}^3 $ are the outputs of the neural network. The NN parameters are first trained using $480$ combinations of initial conditions and source terms offline, and then utilized for online reinforcement control. 
\subsubsection{Results}
\begin{figure}[H]
   \centering
   \includegraphics[width=0.3\textwidth]{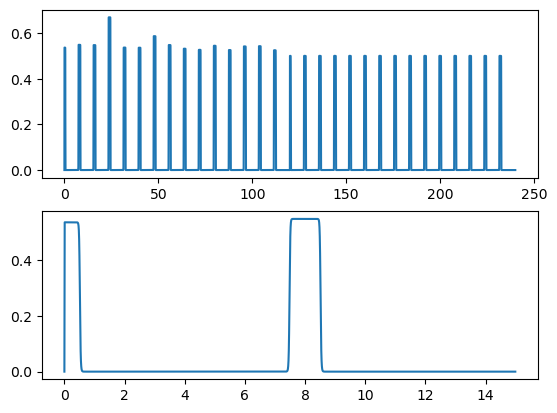}
   \includegraphics[width=0.3\textwidth]{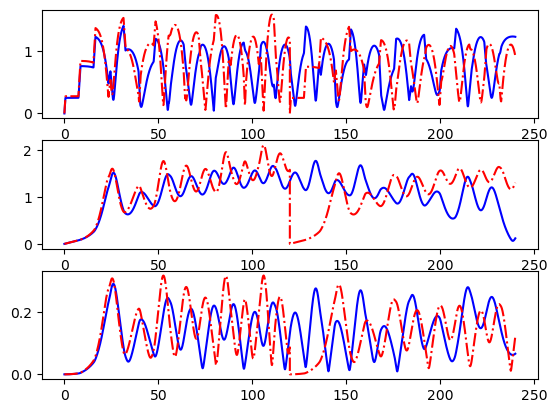}
   \includegraphics[width=0.3\textwidth]{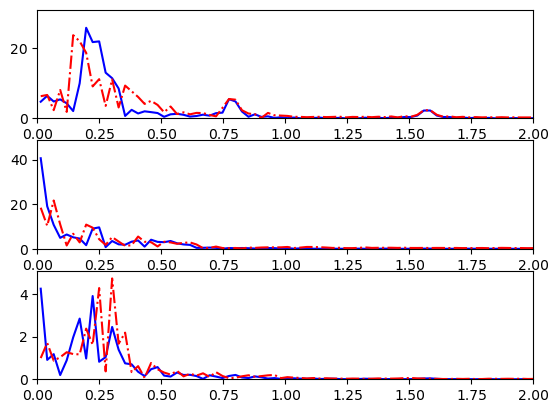}
   \caption{Result for controlling SRS instability: Left column, the optimal learned control policy, and the zoomed-in view of it. The time series and the spectrum of the absolute values of states $a_i$ are shown in the middle and right columns, respectively. Blue lines correspond to our result, and the red lines correspond to the high-fidelity DNS result.}
   \label{fig:srs}
 \end{figure}

The control result for SRS instability is shown in Fig.~\ref{fig:srs}. The first column shows the control signal learned by the hybrid environment and the zoomed-in region of the signal. The optimal policy provides a pulse-like signal to control the spectral properties. The second and third columns show the comparison of the systems response in the state space and spectrum space, respectively. The red line denotes the result obtained by directly applying the control signal to the HF DNS environment, while the blue line shows the result for the hybrid environment. While we cannot precisely match the trajectory, due to the system's chaotic behavior, the spectral properties closely match between two settings, indicating the hybrid environment is an effective surrogate for controlling the high-fidelity environment. More baseline results are shown in Tab.~\ref{tab:SRSTable}, where the hybrid environment performs best among these models. Moreover, we also consider two different settings for the observation variables. In the 1-step case, we only consider the last step in the window as the observation and the the case of time series considers the entire trajectory. Not surprisingly, we observe performance improvements by considering the time history. In the results reported in Table ~\ref{tab:SRSTable}, KL refers to the Kullback–Leibler divergence and SMSE denotes the spectral mean square error measured as the two-norm of the spectral discrepancy.

\begin{table}[htp]
\begin{center}
\setlength\tabcolsep{10pt}
\caption{The evaluation metrics for the Stimulated Raman Scattering (SRS) system, with the HF DNS as the best-possible control result. We compare four different baseline environments. LF ODE is the ODE model for SRS. The hybrid ODE model is the proposed ODE model with learnable terms parameterized by NN, as shown in Eq.~\ref{eq:HyrbidSRS}. Decoder-only Transformer~\cite{han2022predicting} and Siren~\cite{sitzmann2020implicit} are data-driven surrogate environments that we consider as baselines. The 1-step setting uses only the last time step in the time series to compute the reward function, while the time history setting treats the entire time series as the observation. }

\resizebox{\textwidth}{!} 
{\begin{tabular}{ccccccc} 
 \hline
Case setup &&\multicolumn{5}{c}{ \shortstack{Stimulated Raman Scattering (SRS)} }\\
{Env} &{Observations}& $\ell_1$ & $\ell_2$ & $\Delta_{\Omega}$ & \text{KL} & \text{SMSE}\\
 \hline
\multirow{2}{*}{HF DNS} & 1-step & 2.13 & 1.24 & 0.91 &0 &0\\
& Time history & 2.06 & 1.03 & 0.90 &0 &0\\
\hline
 \multirow{2}{*}{LF ODE} & 1-step &36.15  &18.37  &1.62 & $1.6\times10^{-1}$& $4.88$\\
 &Time history&33.26  &16.62  &1.57 &$1.49\times10^{-1}$ &$4.12$\\
 \hline
\multirow{2}{*}{Hybrid ODE (Ours)}&1-step&$\bm{2.60}$&$\bm{1.30}$&$\bm{0.96}$&$\bm{9.1\times10^{-3}}$ &$\bm{0.76}$\\
&Time history&$\bm{2.38}$&$\bm{1.19}$&$\bm{0.94}$&$\bm{9.0\times10^{-3}}$&$\bm{0.72}$\\
\hline
\multirow{2}{*}{Transformer}&1-step&$3.83$&$5.22$&$1.73$&$4.76\times10^{-2}$ &$2.54$\\
&Time history&$3.55$&$4.98$&$1.67$&$4.41\times10^{-2}$&$2.14$\\
\hline
\multirow{2}{*}{Siren}&1-step&$5.28$&$7.44$&$1.40$& $1.33\times10^{-1}$&$4.37$\\
&Time history&$5.08$&$5.61$&$1.38$&$1.30\times10^{-1}$&$4.34$\\
\hline
\end{tabular}}

\label{tab:SRSTable}
\end{center}
\end{table}

\subsection{Fluid Instabilities}
\label{sec:result_fluid}
Next, we proceed to evaluate the effectiveness of our approach in a fluid dynamics use-case. Typically, turbulence occurs when the Reynolds number is high and leads to chaotic behavior and flow instabilities. We use the one-dimensional stochastic Burgers equation (SBE) because of its remarkable ability to mimic fully-developed three-dimensional turbulence flows~\cite{basu2009can}. Moreover, it is a good testbed to benchmark control algorithms for turbulent flow~\cite{choi1993feedback}

The high-fidelity model for stochastic Burgers equation (SBE) of Direct Numerical Simulation (DNS) is given as
\begin{equation}
\label{eq:burgulence}
\frac{\partial{\boldsymbol{u}}}{\partial{t}} +\boldsymbol{u}\frac{\partial{\boldsymbol{u}}}{\partial{\boldsymbol{x}}}=\nu \frac{\partial^2{\boldsymbol{u}}}{\partial{\boldsymbol{x}}^2} +S(\boldsymbol{a}, \boldsymbol{x},t),
\end{equation}
where $\boldsymbol{u}(x,t) = (u_1, u_2, ... ,u_{N_{D}}) \in \mathbb{C}^{N_{D}}$ is the state variable. $\boldsymbol{x}$ is the fine spatial grid and $N_{D}$ is the number of grids. $\nu$ is the constant viscosity and $S$ is the spatio-temporal term accounting for stochasticity. Specifically, $S$ is parameterized by $S(\boldsymbol{x},\boldsymbol{a}, t) = \eta(\boldsymbol{x})S_0(\boldsymbol{a},t)$. $S_0$ is the same Butterworth filter function described in the previous case and $\eta(\boldsymbol{x})$ has the form:
\begin{equation}
\langle \hat{\eta}(\boldsymbol{k}),\hat{\eta}(\boldsymbol{k}^{\prime})\rangle = 2D_0 |\boldsymbol{k}|^{\beta}\delta(\boldsymbol{k}+\boldsymbol{k}^{\prime}),
\end{equation} 
where $\hat{\eta}(\boldsymbol{k})$ is the spatial Fourier transform of the term $\eta(\boldsymbol{k})$. Furthermore, the spectral slope $\beta$ is set to be $-0.75$ to match the complex multi-fractal behavior and the amplitude $D_0$ is set to $1\times10^{-6}$. The HF DNS solution of SBE is simulated using a Fourier collocation code over the domain $L = 2\pi$. We use an explicit second-order Adams-Bashforth scheme to march in time for numerical stability, with $dt = 1\times10^{-4}$. The nonlinear term $\boldsymbol{u}\frac{\partial {\boldsymbol{u}}} {\partial \boldsymbol{x}}$ is computed in conservative form. In this case, the control goal is to minimize the total turbulent kinetic energy in high frequency regimes.
\begin{equation}
\max_{\theta \sim \boldsymbol{a}_{i,\theta} (\boldsymbol{s}_i)}{\sum_{i=1}^{N} \sum_{f>f_\mathrm{thr}}-E_i(f)},
\end{equation} with the corresponding reward function designed as $ r_i = \sum_{f>f_\mathrm{thr}}-E_i(f)$. Here, $E_i(f)$ represents the energy spectrum within the $i^{th}$ control step at given frequency $f$. Finally, $f_{\mathrm{thr}}$ is a hyperparameter representing the frequency threshold, above which the turbulent kinetic energy is minimized. 
\subsubsection{A hybrid differentiable surrogate model for fluids}
The DNS simulation requires fine grids to resolve Kolmogorov scale dynamics, which is usually too expensive, especially for many-query applications like control. Alternatively, large eddy simulation (LES) is an effective technique where large-scale motions are resolved, but subgrid-scale dynamics are modeled using closure terms. The LES equations require an additional term $-\frac{1}{2}\frac{\partial{\boldsymbol{\tau}}}{\partial\boldsymbol{x}}$ to account for the sub-grid dissipation. However, this term is defined on the unresolved quantities and needs to be approximated by different assumptions~\cite{lilly2000meteorological,wong1994comparison}. In our experiment, we use the NN-based hybrid model described in Sec.~\ref{sec:Method} to approximate the sub-grid scale energy dissipation, which can be expressed
\begin{equation}
\label{eq:hybrid_burgers}
\frac{\partial \Tilde{\boldsymbol{u}}}{\partial{t}}+\Tilde{\boldsymbol{u}}\frac{\partial \Tilde{\boldsymbol{u}}}{\partial \boldsymbol{x}}=\nu \frac{\partial^2{\Tilde{\boldsymbol{u}}}}{\partial{\boldsymbol{x}}^2} +S(\boldsymbol{a}, \boldsymbol{x},t)+N_{SGS}.
\end{equation}
Here, $\tilde{\boldsymbol{u}}$ denotes the state variables in the LES model, which are defined on a coarse mesh grid and satisfy the following relation:  $\tilde{\boldsymbol{u}} = G_2(\boldsymbol{u})$. Furthermore, $G_2$ represents the low-pass filter and the cutoff frequency is $M/2$, half the grid number of the hybrid models, and $N_{SGS}$ is the output of the neural network model accounting for the discrepancy between $\boldsymbol{u}$ and $\tilde{\boldsymbol{u}}$. 

\subsubsection{Results}
\begin{figure}[H]
   \centering
   \includegraphics[width=0.4\textwidth]{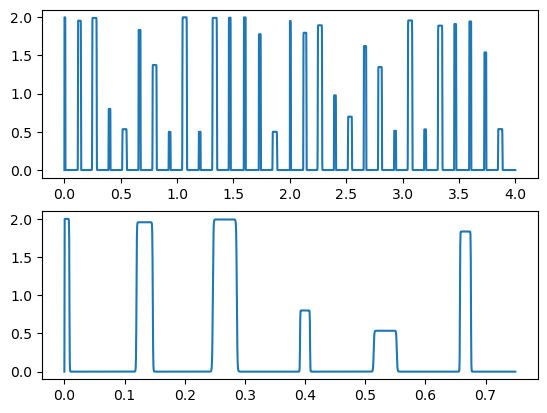}
   \includegraphics[width=0.4\textwidth]{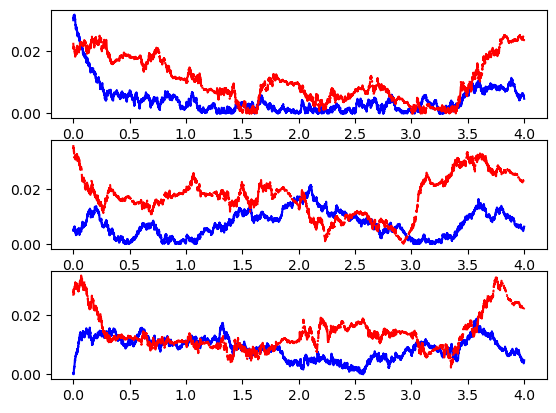}
   \hfill
   \includegraphics[width=0.5\textwidth]{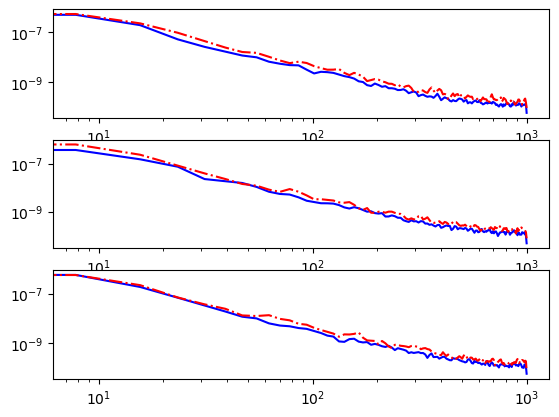}
   \includegraphics[width=0.3\textwidth]{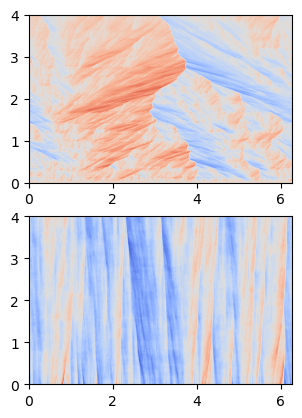}
   
   \caption{Result for controlling SBE instability. The upper left figure shows the optimal control policy and the zoomed-in view. The upper right and lower left figures represent the time series and spectrum of the state variables at locations $i\in[1, M/4, M/2]$, respectively. Finally, the lower right contour compares the uncontrolled and controlled dynamics.}
   \label{fig:sbe}
 \end{figure}

Fig.~\ref{fig:sbe} shows the result for SBE cases.
The optimal policy is trained by a hybrid environment and shown in upper left. The optimal actions has variable periods and amplitudes of the pulse signals. The times series comparison for spatial locations $ i \in [1, M/4, M/2]$ are shown in upper right. The controlled state using HF DNS environment is marked by red, and the one with a hybrid environment is marked by blue. Though they do not visually match in the time domain, the plot in the frequency domain matches well in lower left, where the x-axis $k$ and the y-axis $E(k)$ are presented in the log-log scale. This agreement indicates the statistically close relation between the response generated by the HF and hybrid environments. Qualitatively, the contour of uncontrolled and controlled systems (using Hybrid LES) is plotted in the lower right. The uncontrolled system has a finer structure, while the controlled system looks less chaotic. Quantitatively, the comparison is shown in Tab.~\ref{tab:BurgulenceTable}, where we show metrics for 5 different environments and 2 settings of the observations. Our proposed Hybrid LES environment is the closest result to the HF DNS environment, which is the best possible control we can achieve. Note that there are differences between LF DLES and Hybrid LES models. The DLES is the direct large eddy simulation that solves Eq.~\ref{eq:hybrid_burgers} without NN terms or sub-grid scale modeling. Lacking the energy dissipation model makes the model unstable and blows up in longer-time simulations. 

\begin{table}[htp]
\begin{center}
\setlength\tabcolsep{10pt}
\caption{The evaluation metrics for the SBE system, with the HF DNS control as the best possible control result. We compare four different baseline environments. LF DLES doesn't have any subgrid-scale model. A hybrid model is the LES with learnable terms parameterized by NN.}

\resizebox{\textwidth}{!} 
{\begin{tabular}{ccccccc} 
 \hline
Case setup &&\multicolumn{5}{c}{ \shortstack{Stochastic Burgers Equation (SBE)} }\\
{Env} &{Observations}& \text{Kurtosis}&$\text{Skewness}$ & $E$ &  \text{KL} & \text{SMSE}\\
 \hline
\multirow{2}{*}{HF DNS} & 1-step &$-0.34$& $4.37\times10^{-2}$ & 11.32 & 0 &0\\
& Time history &$-0.33$& $4.02\times10^{-2}$ & 10.96 & 0 & 0\\
\hline
 \multirow{2}{*}{LF DLES} & 1-step &N/A&N/A  &N/A  &N/A & N/A\\
 &Time history&N/A&N/A  &N/A  &N/A &N/A\\
 \hline
\multirow{2}{*}{Hybrid LES (Ours)}&1-step&$\bm{-0.32}$&$\bm{4.76\times10^{-2}}$&$\bm{9.26}$&$\bm{1.50\times10^{-4}}$& $\bm{1.09\times10^{-5}}$\\
&Time history&$\bm{-0.32}$&$\bm{4.20\times10^{-2}}$&$\bm{9.73}$&\bm{$1.37\times10^{-4}$}&$\bm{0.83\times10^{-5}}$\\
\hline
\multirow{2}{*}{Transformer}&1-step&$-0.59$&$9.36\times10^{-3}$&$5.68$&$7.82\times10^{-3}$& $6.15\times10^{-3}$\\
&Time history&$-0.58$&$1.17\times10^{-2}$&$5.61$&$7.68\times10^{-3}$&$6.02\times10^{-3}$\\
\hline
\multirow{2}{*}{Siren}&1-step&$-0.66$&$-1.16\times10^{-1}$&$3.93$&$3.84\times10^{-2}$&$2.67\times10^{-2}$ \\
&Time history&$-0.62$&$1.03\times10^{-3}$&$3.72$&$5.15\times10^{-2}$&$3.41\times10^{-2}$\\
\hline
\end{tabular}}

\label{tab:BurgulenceTable}
\end{center}
\end{table}

\section{Conclusions and limitations}
\label{sec:Conclusions}
We propose a multi-fidelity reinforcement framework for controlling complex dynamical systems. By integrating different hybrid environments into the framework and using spectrum-based reward functions, the control framework can reduce instabilities in physical systems. It outperforms other competitive baseline models and is close to the control directly on the real environments. As for the positive/negative social impact, the current work proposed a framework for benchmark cases in physics. The paper itself aims to incorporate machine learning technique for better control of complex physics systems. It could have positive social impacts that help developing more physics-informed AI models. However, at current stage it doesn't lead to any real-world applications that could have negative social impacts. To the author's best knowledge, the proposed work doesn't have any potential malicious or unintended uses, fairness considerations, privacy consideration or security considerations. Lastly, the current model still has one limitation: hybrid, different models for 3D environments could be costly to get direct training on the high dimensional data. In future work, we will design new learning architectures to overcome this issue.
\section*{Acknowledgements}
This work was performed under the auspices of the U.S. Department of Energy by the Lawrence Livermore National Laboratory under Contract No. DE-AC52-07NA27344.  The work is reviewed and
released under LLNL-CONF-865107.

\bibliographystyle{unsrt}
\bibliography{ref}


\newpage
\appendix

\section{Appendix}
\label{sec:app}

\subsection{Training process}
\begin{minipage}{0.9\textwidth} 
\begin{algorithm}[H]
\caption{Hybrid Environment Training Process}
\label{alg:training_algorithm}
\begin{algorithmic}[1]

    \State {\bfseries Input:}  Corrected  current step state variables $\mathbf{u}_t^{C}$, current step source term $S_0(\mathbf{a}_t, t)$, physical parameters for the system $\mathbf{\mu}$, low fidelity solver $\hat{F}^{L}$, trainable parameter for neural network $\theta$
    \State {\bfseries Output:} Corrected next step state variable $\mathbf{u}_{t+1}^{C}$.
    
  \For{$i \in [1, N]$}
    \State $\mathbf{u}_{t+1}^{'} = \hat{F}^{L}(\boldsymbol{u}_t, \boldsymbol{a}_t; \boldsymbol{\mu})$
    \State $\mathbf{u}_{t+1}^{C} = \mathbf{u}_{t+1}^{'} + f_{\theta}(\boldsymbol{u}^{C}_{t}, \mathbf{u}_{t+1}^{'},\boldsymbol{a_t};\boldsymbol{\mu},\boldsymbol{\theta})$
    \State Compute $\nabla_{\theta} \leftarrow \nabla_{\theta} \mathcal{L}(\theta,\mathbf{u}_{t+1}^{C}, {\bar{\boldsymbol{u}}}^{H}_{i+1})$
    \State Update $\theta$ using the gradients $\nabla_{\theta}$
  \EndFor    
\end{algorithmic}
\end{algorithm}
\end{minipage}

\begin{minipage}{0.9\textwidth} 
\begin{algorithm}[H]
\caption{Twin-delayed Deep Deterministic Policy Gradient}
\label{alg:trainingRL_algorithm}
\begin{algorithmic}

    \State {\bfseries Input:} state variable $\boldsymbol{u}$, action variable $\boldsymbol{a}$. Tranable parameters $\boldsymbol{\theta}_{\pi}$, $\boldsymbol{\theta}_{q_1}$ and $\boldsymbol{\theta}_{q_2}$.\\
    Initialize actor network $\pi(\boldsymbol{u};\boldsymbol{\theta}_{\pi})$, critic networks $q_1(\boldsymbol{u}, \boldsymbol{a};\boldsymbol{\theta}_{q_1})$ and $q_2(\boldsymbol{u}, \boldsymbol{a};\boldsymbol{\theta}_{q_2})$\\
    Initialize $\pi_{targ} =\pi, $ $q_{targ,1} =q_1$ and $q_{targ,2} =q_2$.\\
    
  \For{$t \in [1, N_t]$}
    \State Execute action in $\boldsymbol{a}_i = \pi(\boldsymbol{u}; \boldsymbol{\theta}_\pi)$ in environment $\hat{\boldsymbol{F}}$
    \State Save new data pair ($\boldsymbol{u}_{i}^{o}$, $\boldsymbol{a}_i$, $\boldsymbol{u}_{i+1}^{o}$, $r_i$, $d_i$) to buffer $\mathcal{D}$.
    \If{$\boldsymbol{d}_i = \text{True}$}{ reset $\hat{\boldsymbol{F}}$}
    \EndIf
    \If{$mod(t,n_{\text{update}})=0$}{}
    \EndIf
    \For{$k \in [1, N_{RL}]$}
    \State Sample J state-action pairs $(\boldsymbol{u}_{j}^{o}, \boldsymbol{a}_j, \boldsymbol{u}_{j+1}^{o}, r_j, d_j)$
    \State Compute $\nabla_{\theta_{q_i}} \leftarrow \nabla_{\theta_{q_i}} \frac{1}{J}\sum_{j=1}^{J}[q_i(\boldsymbol{u}_{j}^{o}, \boldsymbol{a}_j)-Q_j]^{2}$ where $Q_j = r_j+\gamma(1-d_j)\min\{q_{targ,i}(\boldsymbol{u}_{j+1}^{o}\,\pi_{targ}(\boldsymbol{u}_{j+1}^{o}))\}$
    \State Update $\boldsymbol{\theta}$ using the gradients $\nabla_{\boldsymbol{\theta}_{q_i}}$
    \If{$mod(k,2)=0$}{\\ \hskip 4em Compute $\nabla_{\theta_{\pi}} \leftarrow \nabla_{\theta_{q_1}} \frac{1}{J}\sum_{j=1}^{J}q_i(\boldsymbol{u}_{j}^{o}, \boldsymbol{a}_j)$ 
    \\ \hskip 4em Update $\boldsymbol{\theta}_{q_{targ},i}
    $ and $\boldsymbol{\theta}_{\pi_{targ}}$
    
    }
    \EndIf
    \EndFor
  \EndFor    
\end{algorithmic}
\end{algorithm}
\end{minipage}

\subsection{Dataset Details}
\label{sec:training_details}
One plasma simulation dataset, stimulated Raman scattering and one fluid simulation dataset, stochastic Burgers equation are used in the experiments. The governing equation are introduced in Sec.~\ref{sec:result_plasma} and Sec.~\ref{sec:result_fluid}. All the simulations are run on Apple M1 Max. And the detailed parameters are listed in Tab~\ref{tab:CFDDatadetail}.

\begin{table}[htp]
\begin{center}
\caption{Simulation details of datasets}
\begin{tabular}{c|c|c|c|c} 
 \hline
 Dataset & Type & \# Grids & \# Steps & Meshing\\  
 \multirow{2}{*}{Stimulated Raman Scattering} & High Fidelity & $100$ & $1200$&Regular Grid\\
  & Low Fidelity & N/A & $600$&N/A\\
 \multirow{2}{*}{Stochastic Burgers Equation} & High Fidelity& $1024$ & $8000$&Regular Grid \\
  & Low Fidelity &$512$ &$4000$ &Regular Grid\\
 \hline
\end{tabular}

\label{tab:CFDDatadetail}
\end{center}
\end{table}

\subsection{Additional details for experimental setups}
We described the details of the experiments of hybrid environment learning and RL optimization. We provide the hyperparameters used in the experiments in 
 Tab~\ref{tab:hyperparmeters}. All experiments are run on Nvidia Tesla V100 with 16 GB memory.
\begin{table}[h]
\centering
\begin{tabular}{lcc}\toprule
                              \multicolumn{1}{c}{}       & SRS       & SBE                 \\\midrule
                              \textbf{Hybrid Env Optimization}\\
                              Learning rate & $1\times10^{-4}$&$1\times10^{-5}$     \\
                              Optimizer &  \multicolumn{2}{c}{Adam~\cite{kingma2014adam}}                                      \\
                              Batch size&\multicolumn{2}{c}{32}\\
                              Number of epochs&  $2000$&$5000$\\\midrule
                              \textbf{RL optimization}\\
                              Algorithm & \multicolumn{2}{c}{TD3} \\
                                Learning rate & $1\times 10^{-3}$&$1\times 10^{-3}$     \\
                              Optimizer &  \multicolumn{2}{c}{Adam}                                      \\
                              Batch size&\multicolumn{2}{c}{128}\\
                              Number of epochs&  $400$&$1000$\\
                              Discount factor&\multicolumn{2}{c}{0.977} \\
                              Policy Action after &100&200 \\
                              \midrule

                               \textbf{Hybrid NN Architecture}\\
                              Layers           & \multicolumn{2}{c}{3}                                              \\
                              Hidden dimension
                              & \multicolumn{2}{c}{256}                                             \\
                              Output dimension&$12$&$1024$                                             \\
                              Activation function         & \multicolumn{2}{c}{sine}                                          \\\midrule
                              \textbf{RL NN Architecture}\\
                              Layers           & \multicolumn{2}{c}{$2$}                                                        \\
                              Hidden dimension& $64$ &$64$
                                \\
                              Activation function & \multicolumn{2}{c}{relu}\\
                              Output Activation function&\multicolumn{2}{c}{tanh}
                                  \\\bottomrule
\end{tabular}
\caption{Hyperparameters}\vspace{-1em}
\label{tab:hyperparmeters}
\end{table}

\end{document}